\documentclass[sigconf, screen]{acmart}
\usepackage{amsmath}

\copyrightyear{2022} 
\acmYear{2022} 
\setcopyright{rightsretained} 
\acmConference[MM '22]{Proceedings of the 30th ACM International Conference on Multimedia}{October 10--14, 2022}{Lisboa, Portugal}
\acmBooktitle{Proceedings of the 30th ACM International Conference on Multimedia (MM '22), October 10--14, 2022, Lisboa, Portugal}\acmDOI{10.1145/3503161.3548050}
\acmISBN{978-1-4503-9203-7/22/10}

\begin{document}

\title{Rail Detection: An Efficient Row-based Network \\ and A New Benchmark}
\renewcommand{\shorttitle}{Rail Detection}
\author{Xinpeng Li}
\affiliation{%
  \institution{Shenzhen Technology University}
  }
\email{li.xin.peng@outlook.com}

\author{Xiaojiang Peng}
\authornote{*Corresponding author.}
\affiliation{%
  \institution{Shenzhen Technology University}
  }
\email{pengxiaojiang@sztu.edu.cn}

\renewcommand{\shortauthors}{Xinpeng Li \& Xiaojiang Peng}

\begin{abstract}
Rail detection, essential for railroad anomaly detection, aims to identify the railroad region in video frames. Although various studies on rail detection exist, neither an open benchmark nor a high-speed network is available in the community, making algorithm comparison and development difficult. Inspired by the growth of lane detection, we propose a rail database and a row-based rail detection method. In detail, we make several contributions: (i) We present a real-world railway dataset, Rail-DB, with 7432 pairs of images and annotations. The images are collected from different situations in lighting, road structures, and views. The rails are labeled with polylines, and the images are categorized into nine scenes. The Rail-DB is expected to facilitate the improvement of rail detection algorithms. (ii) We present an efficient row-based rail detection method, Rail-Net, containing a lightweight convolutional backbone and an anchor classifier. Specifically, we formulate the process of rail detection as a row-based selecting problem. This strategy reduces the computational cost compared to alternative segmentation methods. (iii) We evaluate the Rail-Net on Rail-DB with extensive experiments, including cross-scene settings and network backbones ranging from ResNet to Vision Transformers. Our method achieves promising performance in terms of both speed and accuracy. Notably, a lightweight version could achieve 92.77\% accuracy and 312 frames per second. The Rail-Net outperforms the traditional method by 50.65\% and the segmentation one by 5.86\%. The database and code are available at: \url{https://github.com/Sampson-Lee/Rail-Detection}. 
\end{abstract}




\maketitle

\section{Introduction}

Rail detection is to identify the railroad region within a video frame. Given an input image, an algorithm outputs a segmentation or anchors to denote a railroad region. Recently, train driving safety has attracted increasing attention in the signal processing and multimedia community, including obstacle detection and damage detection \cite{mukojima2016moving, lu2020scueu, wei2020multi}. Identifying the railroad region is vital for the above tasks. The main challenges of rail detection include lighting variety, rail changes, and high-speed requirements \cite{wang2019railnet}.

\begin{figure}[t]
\center
\includegraphics[width=1\linewidth]{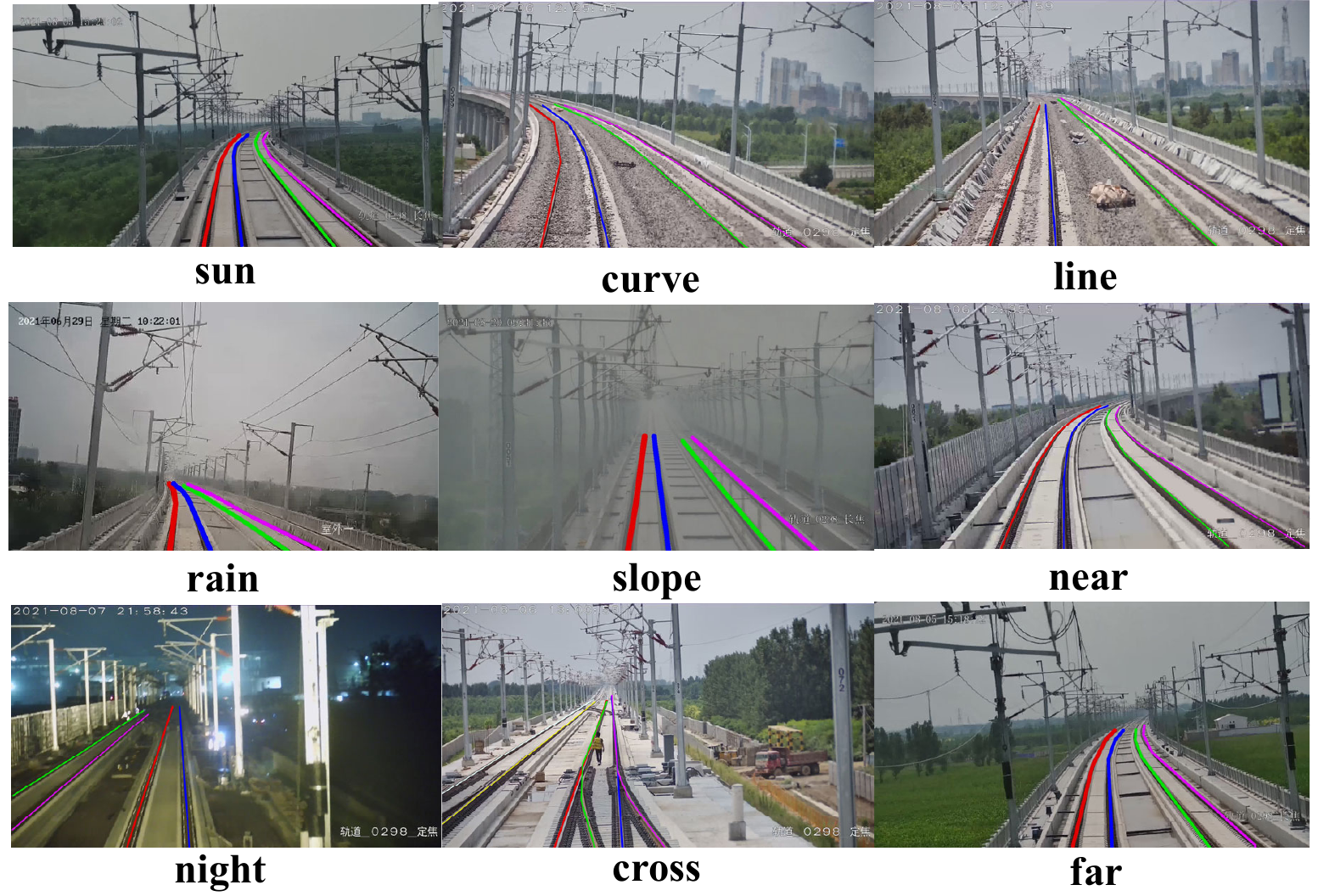}
\caption{Samples on Rail-DB. The rails are labeled with polylines, where colors encode the rails' order numbers. The situations include different lighting, structures, and views: sun, rain, night, curve, slope, cross, line, near, and far.}
\label{fig: image samples}
\end{figure}

In the past decade, various algorithms have been proposed in rail detection, which can be mainly categorized into 1) traditional hand-crafted rail detection methods \cite{kaleli2009vision, nassu2011rail, rodriguez2012obstacle, qi2013efficient, arastounia2015automated, yang2014automated, teng2016visual, 2017Infrastructure}, and 2) deep learning based rail segmentation methods \cite{wang2019railnet, giben2015material}. The traditional hand-crafted rail detection methods usually first extract linear features of an image and then detect rails. These approaches achieve favorable results in simple conditions. Differently, the deep learning based rail segmentation ones adopt sufficient data and convolution neural networks for detection and segmentation. These methods show robustness in real-world environments.

Although various studies on rail detection exist, neither an open benchmark nor a high-speed network is available in the community. It leads to a significant gap in algorithm comparison and development. Recently, the algorithm of lane detection has been booming partly due to the rich and large databases \cite{aly2008real, xu2020curvelane, behrendt2019unsupervised, lee2017vpgnet, pan2018spatial} and high-speed frameworks \cite{yoo2020end, tabelini2021keep, qin2020ultra, tian2018lane}. This paper proposes an open rail database, Rail-DB, inspired by lane detection. The Rail-DB consists of 7432 pairs of images in diverse backgrounds and high-quality annotations. Additionally, we present a row-based network, Rail-Net, to meet high-speed and accuracy requirements.

First, considering the existing rail database \cite{wang2019railnet} is unavailable for further research, we construct the Rail-DB and make it publicly available for the community. To build the Rail-DB, we first obtain videos from real-world trains in different conditions and collect 7432 images. Figure \ref{fig: image samples} illustrates some examples of the Rail-DB. The lighting covers the sun, rain, and night. The structures range in the curve, slope, cross, and line. The views include near and far. Experiments show that images with diverse environments contribute to the algorithm's robustness. Then, we label rails with polylines through coarse and refinement stages to gather high-quality annotations. At the coarse stage, eight annotators use \textit{LabelMe}\footnote{https://github.com/CSAILVision/LabelMeAnnotationTool} to label every rail of an image. However, non-primary rails and railway tails are occasionally missed in the coarse annotations. Therefore, we need a refinement stage to correct the incomplete and inaccurate labels. An expert checks the above annotations and refines them. Besides, we annotate the images with nine scene properties based on backgrounds. Finally, we get the Rail-DB, a middle-size, background-rich, and polyline annotated dataset, which will be publicly available.

Second, we propose the Rail-Net as a rail detection benchmark. Existing works mainly focus on segmentation algorithm \cite{wang2019railnet, giben2015material}, which is computationally inefficient. Recently, row-based methods are widespread in efficient lane detection \cite{tian2018lane, yoo2020end, qin2020ultra,  tabelini2021keep}, which inspires us to extend them to rail detection. The segmentation methods translate the labeled polylines to a multivalued mask. In contrast, we adopt anchors and grindings to further turn the mask into a series of horizontal locations. The locations are intersections between anchors and polylines, which are shrunk into class units by further gridding.
Therefore, the process of rail detection can be treated as a row-based selection problem. We produce a lower prediction number than outputting a segmentation map. Experiments suggest that the row-based selecting algorithm runs faster than existing segmentation methods while keeping higher accuracy.

Third, we perform extensive experiments on Rail-DB and Rail-Net. The results show that our method achieves promising performance in terms of both speed and accuracy. Notably, a lightweight version could even achieve 92.77\% accuracy and 312 frames per second. The Rail-Net outperforms the traditional method by 50.65\% and the segmentation one by 5.86\%. Besides, ablation studies, including cross-scene setting, griding number, metric threshold, rail and lane comparison, and various backbones, justify the effectiveness of every setting. Further, we find the cross structure and the cross-scene adaptation confusing. In the future, we will further solve these challenges and build a larger dataset.

The contributions can be briefly summarized as follows.:
\begin{itemize}
  \item [1)] We propose the first open rail dataset, Rail-DB. The Rail-DB consists of 7432 pairs of multi-scenario images and high-quality annotations, expected to promote rail detection.
  \item [2)] We present a novel high-speed rail detection network, Rail-Net. We are the first to extend row-selecting for rail detection where only segmentation methods are available. The Rail-Net is demonstrated to be efficient and significantly better than the existing segmentation method. 
  \item [3)] We perform extensive experiments on the Rail-DB and the Rail-Net, including the comparison between rail and lane, the comparison between segmentation and classification, and cross-scene adaptation. The proposed method achieves 92.3\% accuracy and 312 frames per second.
\end{itemize}

\section{Related work}

\subsection{Railroad Detection}
The various algorithms in rail detection can be mainly categorized into 1) traditional hand-crafted feature based rail detection methods \cite{kaleli2009vision, nassu2011rail, rodriguez2012obstacle, qi2013efficient, arastounia2015automated, yang2014automated, teng2016visual}, and 2) deep learning based rail segmentation methods \cite{wang2019railnet, giben2015material}. 

Specifically, \cite{giben2015material} performs visual rail inspection using material classification and semantic segmentation with deep convolutional neural networks. \cite{wang2019railnet} builds a private railroad segmentation dataset and uses a pyramid structure to extract multi-scale features.

\subsection{Lane Detection}
Recently, the algorithm of lane detection has been booming partly due to the rich and large databases \cite{aly2008real, xu2020curvelane, behrendt2019unsupervised, lee2017vpgnet, pan2018spatial} and various frameworks \cite{pan2018spatial, hou2019learning, hou2020inter, ghafoorian2018gan, chougule2018reliable, van2019end, philion2019fastdraw, yoo2020end, tabelini2021keep, xu2020curvelane, lee2017vpgnet, qin2020ultra}.

 The Tusimple \footnote{https://github.com/TuSimple/tusimple-benchmark} and CULane \cite{pan2018spatial} are the two most famous datasets of lane datasets, promoting many algorithms in lane detection. Some efficient row-based methods take advantage of the slender characteristics of the rails to reduce computation costs. Specifically, \cite{qin2020ultra} implements a small-sized anchor-based lane marking detection algorithm on ResNet. \cite{tabelini2021keep} proposes an anchor-based feature pooling and a novel attention mechanism.
 
\section{Rail-DB}

Although many methods have been proposed for rail detection in the community, a public railroad dataset is unavailable. Recently, lane detection has been promoted partly due to various datasets, e.g., Tusimple and CULane \citep{pan2018spatial}. A high-quality dataset is expected to contribute to algorithm performance and comparison in the deep learning era. To this end, we construct a high-quality rail dataset for rail detection. We will introduce the establishment of the Rail-DB and the comparison between the Rail-DB and existing datasets.

\subsection{Dataset collection and annotation}

\begin{figure}[ht]
\center
\includegraphics[width=1\linewidth]{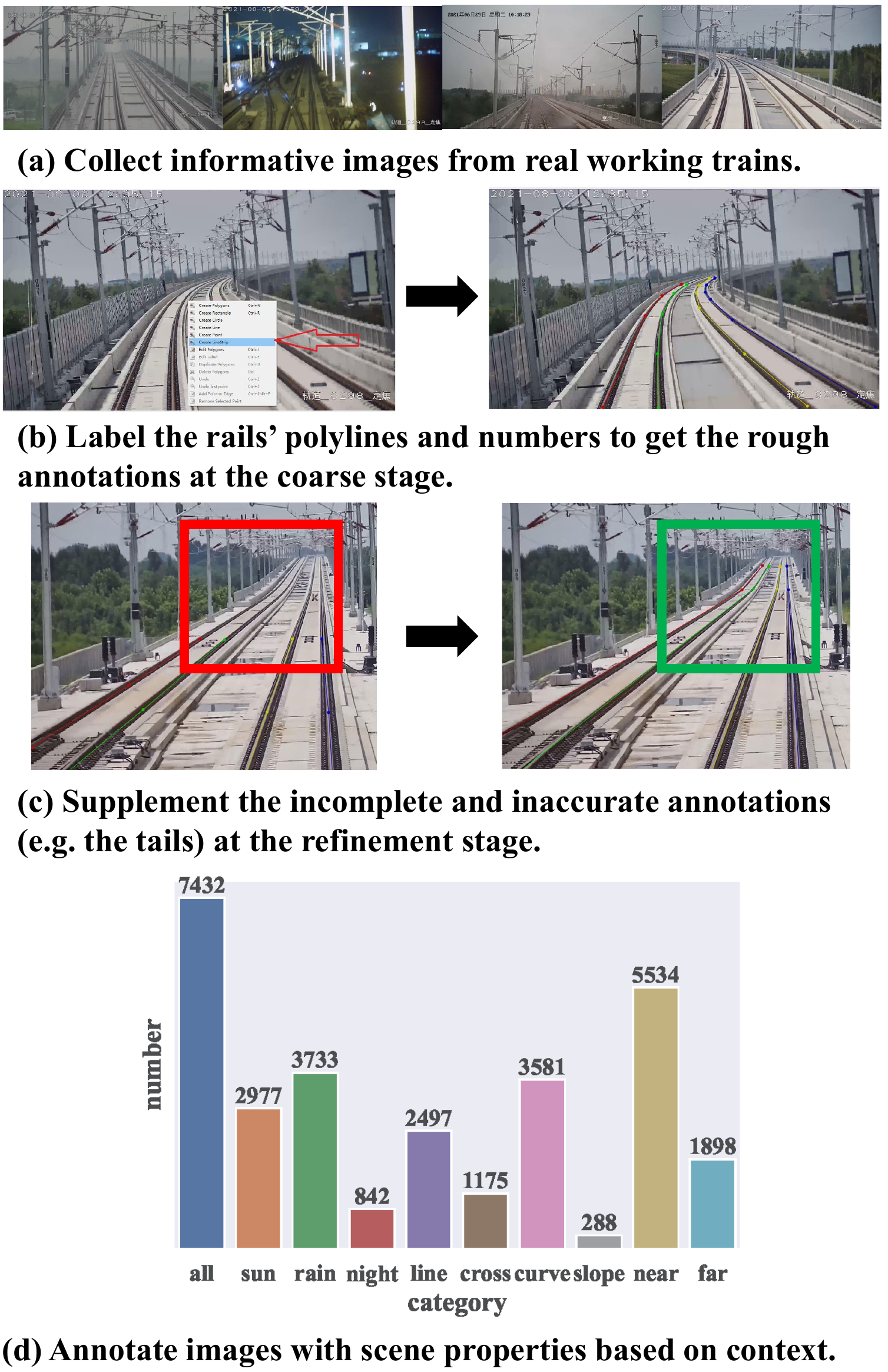}
\caption{The establishment process of our Rail-DB: image collection, coarse annotation stage, refinement annotation stage, and scene annotation. Better to zoom in with PDF.}
\label{fig: dataset collection}
\end{figure}

Figure \ref{fig: dataset collection} exhibits four steps to build the Rail-DB: the image collection stage, the coarse annotation stage, the refinement annotation stage, and the scene annotation stage. 

First, we obtain videos from real-world trains in different conditions to collect rich rail data. Instead of downloading internet images, we capture videos of working trains on various railroads. We use adaptive frame intervals to guarantee sufficient difference between adjacent frames as the train speed varies. We finally got 15 long-term videos and extracted 7432 different frames. Figure \ref{fig: dataset collection} (a) shows some raw frames, which contain varied environments: lighting, road structures, and views. Second, we label rails with polylines at the coarse stage to gather annotations. At the coarse stage, eight annotators use \textit{LabelMe} to label rails' polylines. A polyline represents a rail that fits its varying slender structure. Besides, we annotate different rails in an image with different numbers to encode the trails' priority order. The central pair of rails in an image get numbers 1 and 2 as they are the primary rails of the view. The other rails obtain pairs of increasing numbers according to their growing distance from the main rails. Figure \ref{fig: dataset collection} (b) illustrates the drawing and colored polylines. However, most of the coarse annotations are more or less incomplete and inaccurate in the non-primary trails, the tails, and the priority order. Therefore, we need a refinement stage to improve the quality of the coarse annotations. An expert checks all the rough annotations and refines them carefully. Figure \ref{fig: dataset collection} (c) displays that an incomplete annotation of the tails is made complete at the refinement stage. In addition to the rail annotation, we annotate all images with nine scene properties based on contexts to explore algorithms' generalization. According to various lighting, structures, and views, the contexts are categorized into sun, rain, night, curve, slope, cross, line, near, and far, which account for 2977, 3733, 842, 2497, 1175, 3581, 288, 5534, and 1898 images, respectively. Some images include more than one scene property. Figure \ref{fig: dataset collection} (d) shows the statistics of our Rail-DB.


\subsection{Dataset comparison}

\begin{table*}[ht]
\centering
\begin{tabular}{l|ll|ll}
\hline
Database   & Tusimple                & CULane                  & RSDS & Rail-DB (ours) \\ \hline

quantity     & {\color[HTML]{FFCCC9} } 3626 & {\color[HTML]{FFCCC9} } 133,235 & 3000 & 7432 \\ \cline{1-1}

lighting     & {\color[HTML]{FFCCC9} } sun & {\color[HTML]{FFCCC9} } sun, night, rain & sun & sun, night, rain \\ \cline{1-1}

structures & {\color[HTML]{FFCCC9} } line, curve & {\color[HTML]{FFCCC9} } line, curve, cross & line, curve & line, curve, cross, slope \\ \cline{1-1}

annotation & {\color[HTML]{FFCCC9} } keypoints & {\color[HTML]{FFCCC9} } keypoints & segmentation map & polylines \\ \cline{1-1}

availability  & {\color[HTML]{FFCCC9} } public & {\color[HTML]{FFCCC9} } public & private & public \\ \hline
\end{tabular}
\caption{The comparison of properties between existing lane and rail datasets and the Rail-DB. The Rail-DB is a middle-size, background-rich, polyline annotated, and open rail dataset.}
\label{tab: dataset comparison}
\end{table*}

\begin{figure*}[ht]
\center
\includegraphics[width=0.85\linewidth]{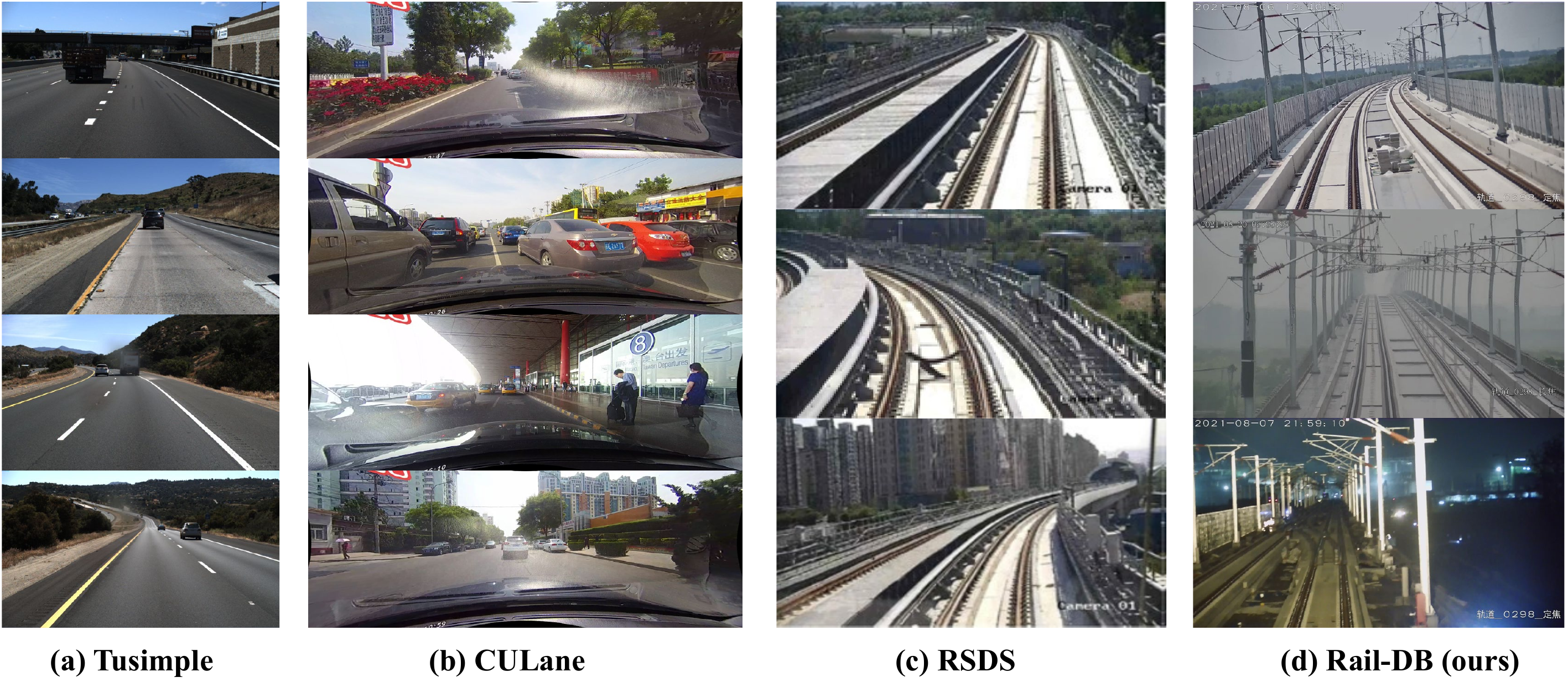}
\caption{The comparison of visualization between the existing database and the Rail-DB. The structures of rails are longer and thinner than lanes. Besides, the view of rails differs from lanes since trains are higher than cars. The images of the RSDS are mainly captured in the city under the sun. In contrast, the images of the Rail-DB are recorded in the wild in richer contexts.}
\label{fig: dataset comparison}
\end{figure*}
We compare the Rail-DB with existing lane and rail datasets in this section. Table \ref{tab: dataset comparison} and Figure \ref{fig: dataset comparison}  show the property and visualization differences between existing datasets and the Rail-DB. The comparison shows the Rail-DB is a middle-size, background-rich, polyline annotated, and open dataset.

The Tusimple \footnote{https://github.com/TuSimple/tusimple-benchmark} and CULane \cite{pan2018spatial} are two famous datasets of lane datasets \cite{aly2008real, xu2020curvelane, behrendt2019unsupervised, lee2017vpgnet, pan2018spatial}. The quantity, lighting, road structures, annotation, and availability are shown in Table \ref{tab: dataset comparison}.
In comparison, rail detection is more challenging than lane detection in three aspects: road structure, complex backgrounds, and dataset availability. From Figure \ref{fig: dataset comparison}, rails are slenderer and more curved than the lane; the rail background contains forests, fields, bridges, city buildings, etc. Worse, there is no public rail dataset. Therefore, a publicly available and multi-scenario rail dataset is urging for rail detection. 


The RSDS \cite{wang2019railnet} is a private rail dataset with 3000 pairs of images and segmentation labels. The lighting contains the sun, and the road structures include lines and curves. The RSDS is private and simple, although this dataset achieves practical results. In comparison, the Rail-DB is more extensive and informative than the RSDS. Specifically, the Rail-DB is an open rail dataset consisting of 7432 pairs of images and polyline labels. The lighting includes sun, night, and rain, and the road types vary in line, curve, cross, and slope. As is shown in Figure \ref{fig: dataset comparison}, the RSDS and the Rail-DB are made in different environments. The images of the RSDS are mainly captured in the city under the sun. In contrast, the pictures of the Rail-DB are recorded in the wild in various environments. 

\section{Rail-Net}

\begin{figure*}
\center
\includegraphics[width=0.9\linewidth]{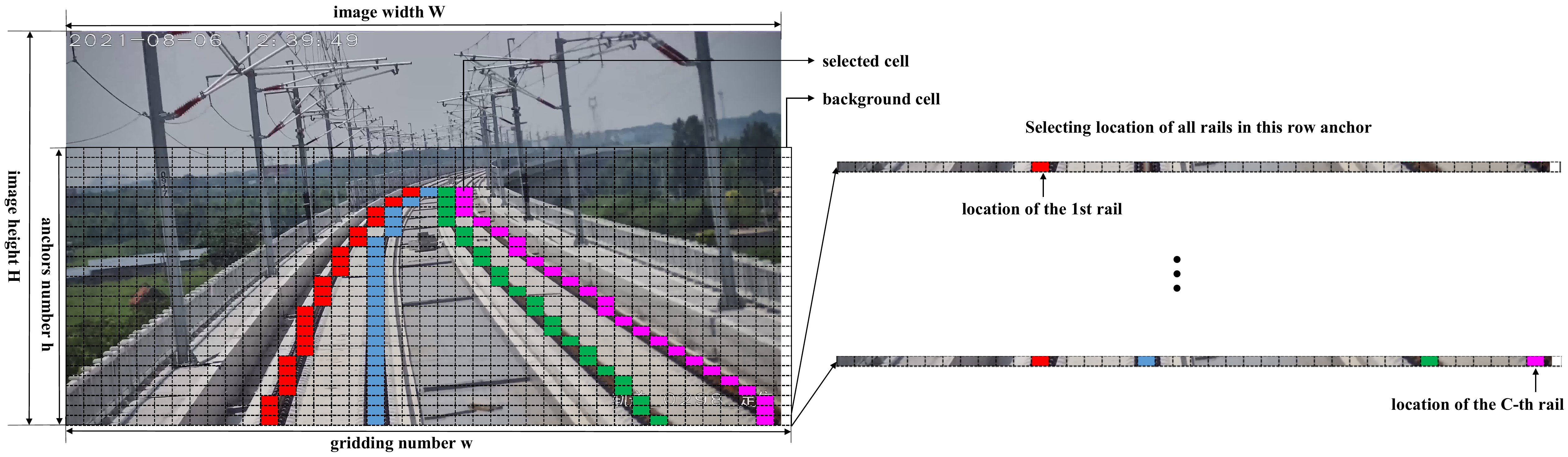}
\caption{Illustration of selecting locations of the rails. Different colors refer to different rails.}
\label{fig: location selection}
\end{figure*}

\begin{figure*}
\center
\includegraphics[width=1\linewidth]{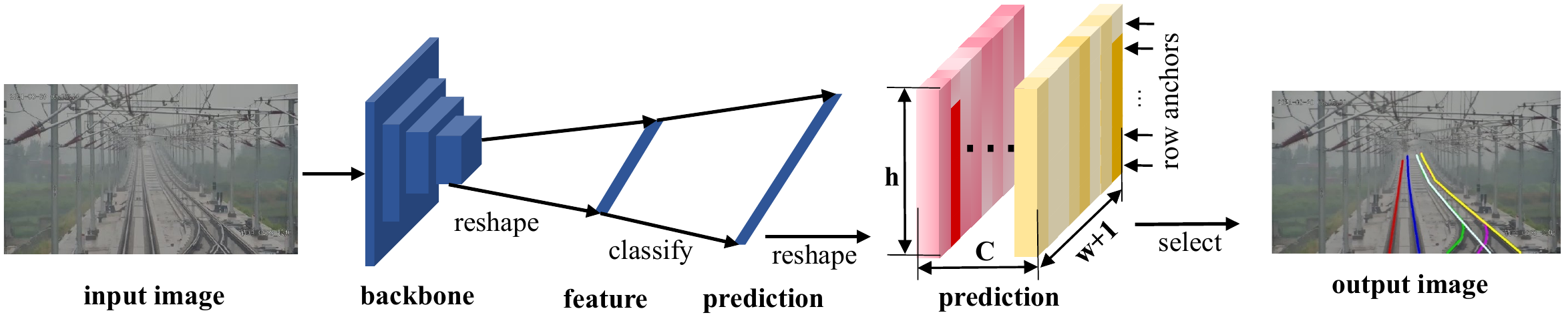}
\caption{The pipeline of our Rail-Net. We first transfer an image into high-level feature maps through a backbone. Then, we reshape the high-level feature maps and classify them for the predictions of all the anchors and rails. }
\label{fig: railnet arch}
\end{figure*}

We propose the Rail-Net to detect the rails in video frames automatically. The existing rail detection works include the traditional hand-craft methods \cite{kaleli2009vision, nassu2011rail, rodriguez2012obstacle, qi2013efficient, arastounia2015automated, yang2014automated, teng2016visual, 2017Infrastructure, le2018railway} and the deep learning segmentation approaches \cite{wang2019railnet, giben2015material}. They are either effective or computation efficient. The efficient row-based methods are widespread in lane detection \cite{yoo2020end, tian2018lane, tabelini2021keep, qin2020ultra}. These methods take advantage of the slender characteristics of the rails to reduce computation costs. To this end, we extend the row-based methods to rail detection. We will introduce the problem formulation and the network pipeline of the Rail-Net in this section.

\subsection{Problem Formulation}

We treat rail detection as a row-based selecting problem instead of a segmentation issue. The segmentation methods turn the polylines into multivalued maps. In contrast, we transfer the labeled polylines of rails into a series of horizontal locations. Figure \ref{fig: location selection} shows selecting areas of the rails in an image.

We adopt the anchors and uniform columns to grid the image into cells. The labeled polylines would intersect with the anchors in specific columns. As shown in Figure \ref{fig: location selection}, we highlight the intersection areas with different colors for different rails, where the corresponding columns with the same colors represent the rails' locations. Therefore, we select these locations as the class labels for each rail. The anchors and the column locations make rail detection a row-based selection problem. Its algorithm outputs location classes, which are less than the one of segmentation pixels.

Suppose the number of rails is $C$, the number of row anchors is $h$, and the number of gridding columns is $w$. Suppose $X$ is the feature extracted through the backbone and $f^{ij}$ is the classifier to predict the rail location on the $i$-th rail, $j$-th row anchor. The formulation of rail prediction can be written as follows:

\begin{equation}
P_{i, j,:} = f^{i j}(X), \text { s.t. } i \in[1, C], j \in[1, h] \text {, }
\label{eq:pred}
\end{equation}
where $P_{i, j,:}$ is the $(w+1)$-dimensional probability vector for selecting gridding columns of the $i$-th rail, $j$-th row anchor. Particularly, we extend the $w$ dimension to $(w + 1)$ dimension to indicate the background. For each rail and each anchor, the algorithm predicts the probability distribution of all locations. As a result, the right locations can be selected based on the probability distribution to represent the complete rails.

The row-based formulation significantly reduces computation costs compared to a segmentation map. The computation cost of our method is proportional to $C \times h \times (w+1)$. Suppose the image size is $H \times W$, and the computation cost of the segmentation method is proportional to $C \times H \times W$. In general, the numbers of gridding cells and row anchors are far smaller than the image size. That is to say, $h<<H$ and $w<<W$. In formulation, the computation reduction ratio ($r$) can be expressed as:
\begin{equation}
\label{eq:ratio}
\end{equation}

Take the setting of the Rail-DB as an example, the input size of the image is $288 \times 800$, and the size of gridding cells is $52 \times 200$. The computation reduction ratio achieves $22$.

\subsection{Network Pipeline}

We propose the Rail-Net based on the row-based formulation. We discard up-sampling modules compared to the segmentation methods. Figure \ref{fig: railnet arch} shows the pipeline of the Rail-Net.

We first extract the high-level feature map for a given image through a backbone. Specifically, an 18-layer ResNet \cite{he2016deep} serves as the backbone for its computation efficiency and performance effectiveness. The input size is $288 \times 800 \times 3$ in the Rail-DB, the output size is $9 \times 25 \times 512$ after a $1/32$ down-sampling in the backbone. 

We classify the high-level feature for the predictions of all the anchors and rails. In detail, a $1 \times 1$ kernel convolutional layer reduces the high-level feature into $9 \times 25 \times 8$. Then a reshape operation flattens the high-level feature into a $1800$ dimensions vector. These operations are more advanced than the global average pooling. The vector contains global spatial information and avoids computation redundancy, meeting the accuracy and high-speed requirements. Finally, a classifier turns the vector into the $C \times h \times (w+1)$ predictions. The classifier consists of an $1800-2048$ linear layer, a ReLU layer, and a $2048-C \times h \times (w+1)$ linear layer. In the setting of the Rail-DB, $C$ is 4, $h$ is 52, and $w$ is 200.

We adopt a Cross-Entropy (CE) function for classification loss $L_{cls}$ to regularize the parameters optimization in the training phase. Suppose $T_{i, j, :}$ is the one-hot label of correct locations. The formulation corresponds to:

\begin{equation}
L_{c l s}=\sum_{i=1}^{C} \sum_{j=1}^{h} {C E}\left(P_{i, j, :}, T_{i, j, :}\right) \text {. }
\end{equation}

We use the expectation of predictions like \cite{qin2020ultra} to obtain locations from the classification prediction in the testing phase. A softmax function gives the probability of different locations:
\begin{equation}
{Prob}_{i, j,:}={softmax}\left(P_{i, j, 1: w}\right) \text {, }
\end{equation}
where $P_{i, j, 1: w}$ is a w-dimensional vector since the background gridding column is not included, and ${Prob}_{i, j,:}$ represents the normalized probability at each location. Then, the formulation of expected locations ${Loc}_{i, j}$ can be written as:
\begin{equation}
{Loc}_{i, j}=\sum_{k=1}^{w} k \cdot {Prob}_{i, j, k} \text {, }
\end{equation}
in which ${Prob}_{i, j, k}$ is the probability of the $i$-th rail, the $j$-th row, and the $k$-th location.

\section{Experiments}


\subsection{Implementations}

\textbf{Dataset splits.} We randomly sample 80\% images as the training set and 20\% as the validation set. We resize images into $1280 \times 720$ in the evaluation following the Tusimple \footnote{https://github.com/TuSimple/tusimple-benchmark}. The predefined row anchors range from 200 to 720 with a step of 10.

\textbf{Training Details.} The optimizer is Adam. The learning rate strategy is cosine decay. The initialized learning rate is 4e-4. The batch size is 64. The total training epochs number is 50. We train and test all models with PyTorch \cite{paszke2019pytorch} and NVIDIA Tesla V100 GPU.

\textbf{Data Pre-process.} In the optimizing process, images are resized to $800 \times 288$ following \cite{pan2018spatial}. A row-based classification network is vulnerable to overfit the training set because of the inherent structure of rails. Therefore, we augment the images with rotation and vertical and horizontal shifts. 

\textbf{Evaluation Metric.} The evaluation metric is accuracy, which is

\begin{equation}
\text { accuracy }=\frac{\sum_{\text {clip }} C_{\text {clip }}}{\sum_{\text {clip }} S_{\text {clip }}},
\end{equation}
where $C_{\text {clip }}$ is the number of rail points predicted correctly and $S_{\text {clip }}$ is the total number of ground truths in each image. If the difference width between ground truth and prediction is less than a threshold, the predicted point is a correct one. We set the threshold $T_p$ by default to 6 pixels for the Rail-DB.

\subsection{Method Comparison}

\begin{table}[]
\begin{tabular}{c|ccc}
\hline
Accuracy (\%) & Hand-crafted & Segmentation   & Rail-Net \\ \hline
sun    & 26.61 & 70.53   & \textbf{88.31}  \\
rain   & 56.55 & \textbf{95.62} & 95.42    \\
night  & 31.09 & 76.61   & \textbf{96.22}  \\
line   & 43.57 & 87.97   & \textbf{95.82}  \\
cross  & 29.88 & 80.15   & \textbf{87.74}  \\
curve  & 45.58 & 88.85   & \textbf{92.76}  \\
slope  & 27.96 & 86.78   & \textbf{91.48}  \\
near   & 44.01 & 84.91   & \textbf{93.75}  \\
far    & 36.57 & \textbf{92.93} & 89.87    \\ \hline
total  & 42.14 & 86.91   & \textbf{92.77}  \\ \hline \hline
FLOPs (G)     & -     & 13.02   & \textbf{8.42}   \\
memory (MB)   & -     & 258.62  & \textbf{182.84} \\ \hline
FPS    & 43    & 188     & \textbf{312}    \\ \hline
\end{tabular}
\caption{Quantitative comparison of all methods on the Rail-DB. In accuracy, the Rail-Net outperforms the hand-crafted method and the segmentation approach by 50.65\% and 5.86\%. In FPS, the Rail-Net runs 7.16 times and 1.64 times faster than the hand-crafted method and the segmentation approach.}
\label{tab: quantitative results}
\end{table}

\begin{figure}[t]
\center
\includegraphics[width=0.9\linewidth]{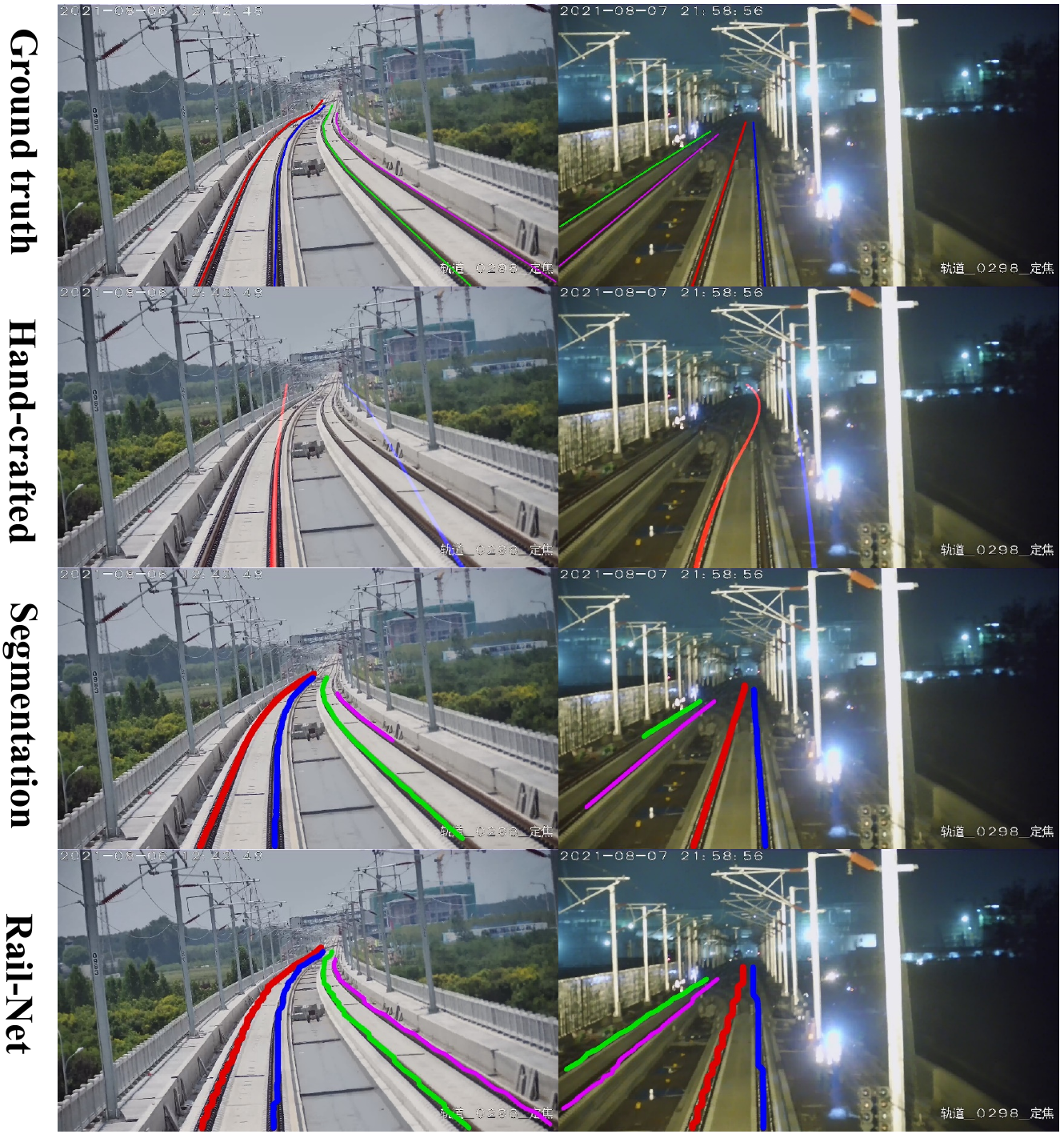}
\caption{Qualitative comparison of all methods on the Rail-DB. Different colors refer to different rails. The Rail-Net is more accurate and complete in the sun and night situations than alternative methods.}
\label{fig: qualitative comparison}
\end{figure}

We compare the Rail-Net with the traditional hand-crafted method and the deep learning segmentation approach on the Rail-DB in this section. The Rail-Net utilizes an 18-layer Resnet \cite{he2016deep} for the backbone and fully linear layers for the classifier. The hand-crafted method uses the Sobel operator \cite{kanopoulos1988design} and polynomial fitting to match railroad rails. The segmentation method selects the same backbone as the Rail-Net, and convolutional blocks for the segmentation heads like FPN \cite{lin2017feature}. We show quantitative and qualitative comparison results in the following.

\textbf{Quantitative comparison.} We compare all methods in the evaluation accuracy and the frame per second (FPS) in these experiments. The validation accuracy of one run is reported as the evaluation accuracy. The FPS is recorded with the average time for 100 runs. We also calculate each situation's accuracy and adopt torchinfo \footnote{https://github.com/TylerYep/torchinfo} to get inference memory and floating point operations per second (FLOPs). The results are shown in Table \ref{tab: quantitative results}.

From Table \ref{tab: quantitative results}, we can see that the Rail-Net achieves dominant performance in accuracy and FPS. For accuracy, the Rail-Net outperforms the hand-crafted method and the segmentation approach by 50.65\% and 5.86\%. For the FPS, the Rail-Net runs 7.16 times and 1.64 times faster than the hand-crafted method and the segmentation approach. The results show the Rail-Net exceeds alternative methods in most situations as the formulation of the Rail-Net is more suitable for slender rails.

However, the segmentation approach outperforms the Rail-Net under the far scenes. Intuitively, the Rail-Net uses the slender property of rails and the contrast of an image row. In most cases, it is more certain and distinctive to output the rails' positions. Nevertheless, the rail areas are more apparent in the far situation due to the enlarged perspective. It is better to segment the big area than locate its location. Additionally, the Rail-Net occupies less computational cost in FLOPs and memory than the segmentation approach, as the row-based formulation replaces segmenting each pixel with selecting the locations of each row. The reduced FLOPs and memory contribute to higher FPS.

\textbf{Qualitative comparison.} Figure \ref{fig: qualitative comparison} shows the visualization of all methods. We color the labels of the ground truth and the predictions under the sun and night conditions. We can see the Rail-Net performs better than other methods. The Rail-Net shows advanced completeness and accuracy in two situations. Specifically, the hand-crafted method fails to recognize the lane class or the distant trail position. The Sobel operator fails to process complicated contexts, leading to odd-fitting curves. Besides, the segmentation methods are confused with the non-primary rails and the background because the secondary rails are much smaller than the main rails. In contrast, the row-based Rail-Net should select a specific location in one row, accounting for more complete results. 

Worth mentioning that the locations seem still precise when the rail extends far ahead since the line becomes thinner. On the one hand, we split the width of 1280 into 200 columns with padding for rail detection, i.e., the horizontal resolution for our method is 6 pixels. It is enough in practical applications, and we do not observe serious degradation in far rail locations. On the other hand, the far rail locations can be accurate in the following frames.

\begin{table*}[]
\begin{tabular}{c|cccccccccc}
\hline
accuracy (\%) & total & sun & rain & night & line & cross & curve & slope & near & far   \\ \hline
total        & \textbf{92.77} & 56.31 & 60.75          & 36.14 & 64.13 & 55.22 & 79.53 & 36.56 & 78.11          & 48.32 \\
sun        & \textbf{88.31}          & 79.61 & 24.66          & 25.47 & 45.54 & 42.84 & 73.12 & 23.64 & 66.93          & 44.56 \\
rain       & 95.42          & 41.35 & \textbf{95.84} & 36.86 & 73.16 & 60.81 & 92.56 & 49.61 & 83.39          & 54.58 \\
night      & \textbf{96.22} & 37.77 & 25.96          & 75.8  & 91.14 & 75.66 & 35.02 & 20.91 & 92.25          & 33.78 \\
line       & \textbf{95.82} & 54.54 & 63.09          & 42.62 & 95.52 & 57.61 & 71.24 & 35.61 & 76.96          & 54.01 \\
cross      & \textbf{87.74} & 58.67 & 34.16          & 33.41 & 52.59 & 80.14 & 41.69 & 20.75 & 80.93          & 32.51 \\
curve      & \textbf{92.76} & 56.11 & 68.94          & 33.21 & 48.22 & 46.69 & 90.79 & 41.65 & 78.62          & 49.77 \\
slope      & \textbf{91.48} & 39.75 & 75.46          & 22.72 & 53.72 & 43.44 & 77.37 & 77.16 & 43.04          & 73.18 \\
near       & 93.75          & 54.19 & 61.42          & 37.77 & 64.21 & 57.77 & 77.55 & 35.65 & \textbf{93.86} & 34.62 \\
far        & \textbf{89.87} & 62.58 & 58.79          & 31.26 & 63.91 & 47.61 & 85.48 & 39.26 & 31.06          & 89.21  \\ \hline
\end{tabular}
\caption{The results of cross-scene experiments. The column keys and the row keys denote the training rail situations and testing rail situations. The training of the whole dataset exceeds the training of the partial one in most cases.}
\label{tab: cross scene}
\end{table*}

\begin{figure*}
\center
\includegraphics[width=0.8\linewidth]{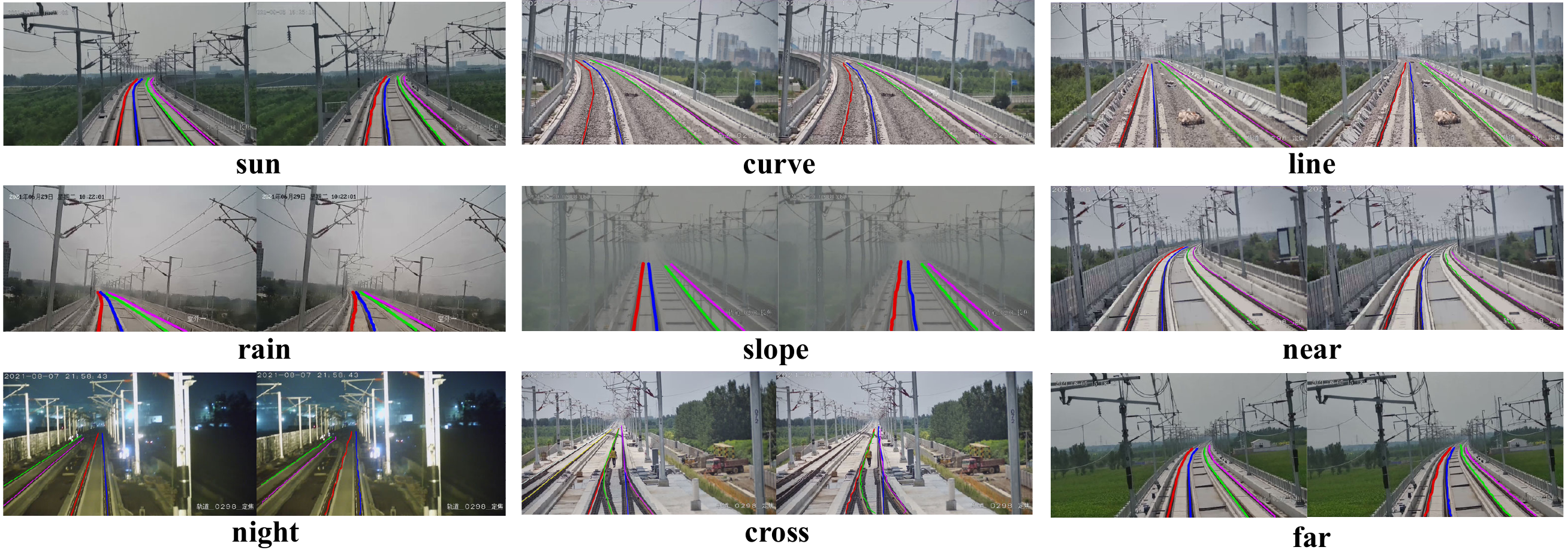}
\caption{The visualization of the predictions under various scenes. The algorithm performs well under different conditions, even in the cross structure. This shows the situation diversity of the images enhances the method's stability.}
\label{fig: cross-scene results}
\end{figure*}

\begin{figure}[ht]
\center
\includegraphics[width=1\linewidth]{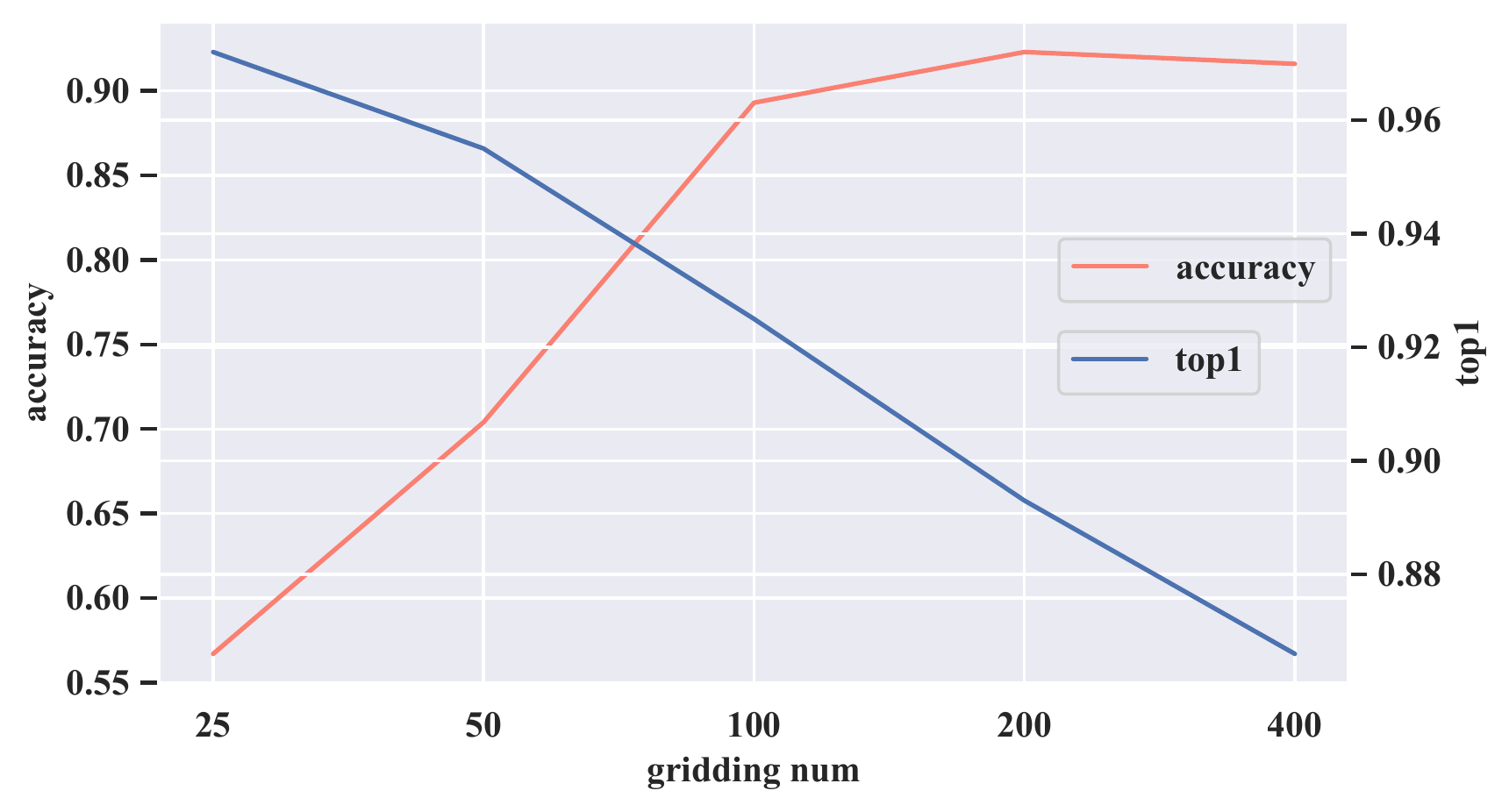}
\caption{The results of different gridding numbers experiments. The evaluation accuracy arrives at the peak when the gridding number is 200.}
\label{fig: res gridding number}
\end{figure}

\begin{figure}[ht]
\center
\includegraphics[width=1\linewidth]{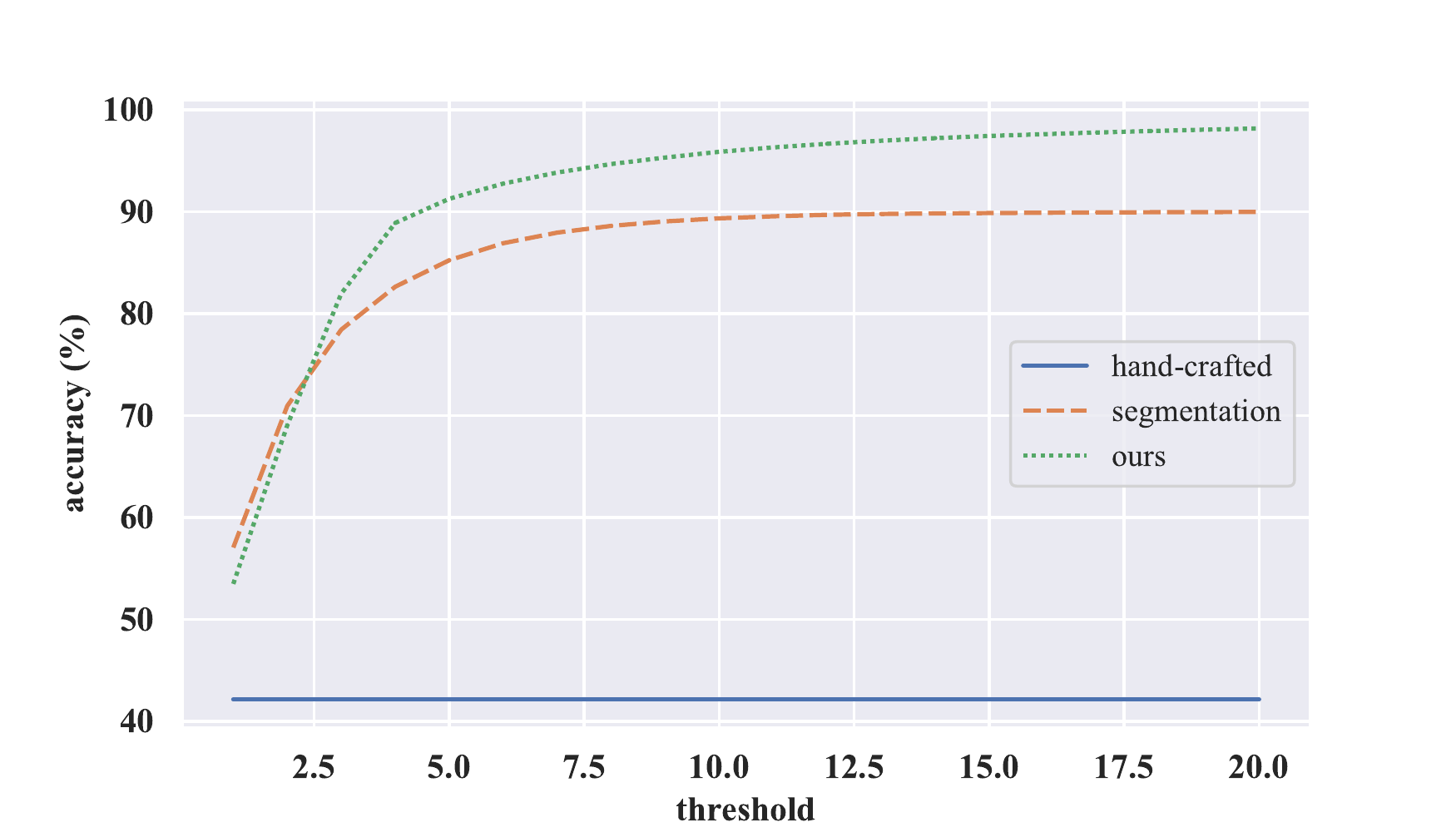}
\caption{The results of different thresholds in metrics. The Rail-Net consistently outperforms the segmentation method with a threshold larger than 3.}
\label{fig: res thresh}
\end{figure}

\begin{table}[h]
\begin{tabular}{l|cc}
\hline
backbone    & accuracy (\%) & FPS \\ \hline
ResNet (18) & 92.77 & 312 \\
ResNet (34) & \textbf{93.09} & 169 \\
ResNet (50) & 92.71 & 126 \\
SqueezeNet  & 90.99 & \textbf{370} \\
MobileNet   & 92.24 & 147 \\
ViT         & 82.41 & 109 \\ \hline
\end{tabular}
\caption{The results of different backbones for our Rail-Net. The ResNet (18) is sufficient for rail detection compared to deeper ResNets, efficient alternative networks, and the ViT.}
\label{tab: res backbone}
\vspace{-1.5em}
\end{table}

\begin{table}[]
\begin{tabular}{l|ccc}
\hline
                  & Ultra-Fast     & LaneATT        & Rail-Net      \\ \hline
Tusimple (Acc \%) & \textbf{95.82} & 95.57          & 95.64         \\
CULane (F1 \%)    & 68.40          & \textbf{75.09} & 68.1          \\
Rail-DB (Acc \%)  & 84.5           & 85.6           & \textbf{92.7} \\ \hline
\end{tabular}

\caption{The comparison of rail and lane detection. The results show it is not trivial to apply the row-selecting idea of lane detection to rail detection since those SOTA methods based on the idea are largely inferior to ours.}
\label{tab: rail and lane}
\vspace{-1.5em}
\end{table}

\subsection{Ablation study}

We conduct ablation studies for our Rail-Net on the Rail-DB to verify the effectiveness of every setting. We will show the effects of scene diversity, gridding number, backbone models, and thresholds.

\textbf{The scene diversity of the Rail-DB.} As described in Section 3.1, we collect railroad images from different conditions to improve scene diversity. To survey the effects of scene variety, we carry out cross-scene experiments among nine situations. We adopt the partial dataset of one scene for training. The quantitative and qualitative results are shown in Table \ref{tab: cross scene} and Figure \ref{fig: cross-scene results}.

From Table \ref{tab: cross scene}, the training of the whole dataset exceeds the training of the partial one in most cases except the rain and near scenes, as the diversity of training images often contributes to higher accuracy for one situation and better generalization. It demonstrates that the Rail-DB with diversified situations is more advanced than the existing single-scene rail dataset. Figure \ref{fig: cross-scene results} shows the algorithm performs well under different conditions, even in the cross structure, because the situation diversity enhances the method's stability.

In addition, the accuracy suffers from a sharp degradation for cross-scene conditions. For cross-lighting, the performance falls dramatically, e.g., the accuracy suffers a 54.14\% decline from the sun to the night scenes. The performance also has a rapid cut for cross-structure, e.g., the accuracy decreases by 37.91\% from the line to the cross. For cross-view, the performance degrades seriously, e.g., the accuracy gets a 58.15\% reduction from the far to the near. The algorithm is trained to distinguish the difference between the original rails and backgrounds. Intuitively, the performance drops when adopting other structures, lighting, and views. In the future, we will pay more attention to solving this challenge.

\textbf{The gridding number of the Rail-Net.} As described in Section 4.1, we adopt gridding to establish classification-based formulation. We make the gridding numbers 25, 50, 100, 200, and 400 to explore the effects of different gridding numbers. Figure \ref{fig: res gridding number} shows the results of the top1 accuracy and the evaluation accuracy. Specifically, the top 1 accuracy refers to the classification accuracy of gridding cells, while the evaluation accuracy means the location accuracy of rails.

From Figure \ref{fig: res gridding number}, the top-1 accuracy decreases as the gridding number increases. Because the more gridding cells, the higher the requirement of the fine-grained classification. However, the evaluation accuracy gets a lift and a slight drop instead of suffering from a monotonic decrease. It is because the gridding number is a trade-off between classification and location. When the gridding number is too small, the gridding cell is too large to represent a precise location. When the gridding number is too big, the classification is too difficult, leading to a slight drop in the evaluation accuracy. The evaluation accuracy arrives at the peak when the gridding number is 200. Therefore, we choose 200 as the default gridding number.

\textbf{The thresholds of metric}. As described in Section 5.1, the performance is closely relevant to the threshold $T_p$ of the metric. Figure \ref{fig: res thresh} illustrates the results in the experiment of Table \ref{tab: quantitative results} with varied $T_p$. The Rail-Net consistently outperforms the segmentation method with a threshold larger than 3. The hand-crafted method maintains the same performance since it only provides curves. However, the segmentation method is superior to our Rail-Net with small thresholds. Intuitively, the gridding of the Rail-Net results in coarse grain outputs, while segmentation provides pixel-wise regions. Worth noting that a threshold like 6 makes sense in practice.

\textbf{The comparison between rail and lane detection}. Table \ref{tab: rail and lane} conduct experiments on lane and rail datasets. a) We adopt the Rail-Net on two public lane datasets, TuSimple and CULane, and respectively achieves 95.64 in accuracy and 68.1 in F1. The Rail-Net is slightly inferior to lane detection methods like Ultra-Fast and LaneATT on lane datasets. b) We adopt Ultra-Fast \cite{qin2020ultra} and LaneATT \cite{tabelini2021keep} for rail detection on the Rail-DB. The Ultra-Fast achieves 84.5, and the LaneATT obtains 85.6, which is much lower than the one of our Rail-Net (i.e., 92.7). These results indicate that a straightforward transfer from the lane detection method to rail detection leads to degradation due to the inherent difference between the two tasks. On the contrary, our elaborately designed yet simple row-selecting Rail-Net achieves the best performance.

\textbf{The backbones of Rail-Net}. As described in Section 4.2, we adopt an 18-layer ResNet as the backbone. We evaluate various backbones for our Rail-Net to show their effects, including ResNet \cite{he2016deep} with different layers, SqueezeNet \cite{iandola2016squeezenet}, MobileNet \cite{howard2017mobilenets}, and Vision Transformer (ViT) \cite{dosovitskiy2020image}. Table \ref{tab: res backbone} shows the accuracy and FPS results of different backbones. In accuracy, all the ResNet models perform similarly, indicating that a small network like ResNet18 is sufficient for the rail feature extraction. SqueezeNet runs faster than other networks but slightly degrades accuracy as a lighter computation leads to a weaker performance. ViT performs worse than the convolutional networks, partly indicating that the square patching of the picture is unsuitable for slender rails.

\section{Conclusion}
We have proposed an open benchmark, Rail-DB, and an efficient row-based framework, Rail-Net, for rail detection. Experiments show that a lightweight version of the Rail-Net achieves 312 frames per second and 92.77\% accuracy on the Rail-DB. 

The contributions are as follows: (i) We propose a real-world railway dataset with 7432 pairs of informative images and high-quality annotations. The Rail-DB is expected to facilitate the improvement of rail detection algorithms. (ii) We present a row-based rail detection framework containing a lightweight convolutional backbone and an anchor classifier. Our formulation reduces the computational cost compared to alternative segmentation methods. (iii) We conduct extensive experiments on the Rail-DB, including cross-scene settings and network backbones ranging from ResNet to ViT.

In the future, we will pay more attention to solving these challenges: confusing situations (e.g., the night lighting and crossroad structure) and cross-scene generalization. Besides, considering the slender coherence of rows for smoother outputs and combining segmentation for more fine-grain results are potential directions. 
\\

\textbf{Acknowledgement}. This work is partially supported by the National Natural Science Foundation of China (62176165), and the Natural Science Foundation of Top Talent of SZTU (grant no.GDRC202131). Many thanks to all the annotators: Shuyi Mao, Junyao Huang, Xubin Huang, Shaoshuo Liu, Zhengchun Yang, Shaolong Zhao, Ping Huang, and Bin Lin.

\bibliographystyle{ACM-Reference-Format}
\balance
\bibliography{main}

\end{document}